\documentclass[10pt, a4paper, fleqn]{article}
\usepackage{lrec2006}
\usepackage{graphicx}
\usepackage{url}
\expandafter\def\expandafter\UrlBreaks\expandafter{\UrlBreaks
  \do\a\do\b\do\c\do\d\do\e\do\f\do\g\do\h\do\i\do\j%
  \do\k\do\l\do\m\do\n\do\o\do\p\do\q\do\r\do\s\do\t%
  \do\u\do\v\do\w\do\x\do\y\do\z\do\A\do\B\do\C\do\D%
  \do\E\do\F\do\G\do\H\do\I\do\J\do\K\do\L\do\M\do\N%
  \do\O\do\P\do\Q\do\R\do\S\do\T\do\U\do\V\do\W\do\X%
  \do\Y\do\Z}

\usepackage{hyperref}
\usepackage{amsmath}

\usepackage{array}
\newcolumntype{P}[1]{>{\centering\arraybackslash}p{#1}}

\title{A Tweet Dataset Annotated for Named Entity Recognition and Stance Detection}

\name{Dilek K\"u\c{c}\"uk$^{\star}$, Fazli Can$^{\ast}$}

\address{$^{\star}$Electrical Power Technologies Department\\
T\"UB\.ITAK Energy Institute\\
Ankara{--}Turkey\\
dilek.kucuk@tubitak.gov.tr\\\\
$^{\ast}$Bilkent Information Retrieval Group\\
Computer Engineering Department\\
Bilkent University\\
Ankara{--}Turkey\\
canf@cs.bilkent.edu.tr\\}

\abstract{Annotated datasets in different domains are critical for many supervised learning-based solutions to related problems and for the evaluation of the proposed solutions. Topics in natural language processing (NLP) similarly require annotated datasets to be used for such purposes. In this paper, we target at two NLP problems, named entity recognition and stance detection, and present the details of a tweet dataset in Turkish annotated for named entity and stance information. Within the course of the current study, both the named entity and stance annotations of the included tweets are made publicly available, although previously the dataset has been publicly shared with stance annotations only. We believe that this dataset will be useful for uncovering the possible relationships between named entity recognition and stance detection in tweets. \\ \newline }

\begin{document}

\maketitleabstract

\section{Introduction}\label{sec:intro}

Language resources are critical, firstly, for building natural language processing (NLP) systems\slash approaches and, secondly, for the evaluation and comparison of these systems\slash approaches. Annotated datasets for different NLP topics are among these language resources. Especially for newly-emerging research topics, the availability of (preferably multilingual) datasets leads to early achievement and application of plausible results. On the other hand, recent trend in a large volume of NLP studies is to find solutions applicable to social media texts like microblogs. In some of these studies on social media analysis, the main focus is traditional problems that have been studied for decades on well-formed texts, while in other studies, new problems mostly peculiar to social media posts are considered. NLP problems like part-of-speech tagging, summarization, and information extraction belong to the first group of studies while topics like tweet normalization belong to the second group.\\

In this paper, we present a tweet dataset annotated with the included named entities and stance information. Named entity recognition (NER) is a well-studied NLP topic for different languages, domains, and text genres (including social media posts). It is usually defined as the extraction and classification of named entities such as person, location, and organization names from texts. On the other hand, stance detection aims to determine the position (usually as favor, against, and neither) of the text owner towards a target of interest \cite{mohammad2016semeval}. It is a research topic considerably more recent than NER.\\

The tweet dataset described in the current paper and its preceding smaller version have been previously used in the stance detection experiments described in \cite{kucuk_can_2018stance} and \cite{kucuk2017joint_stance_ner}, respectively. It is reported in both studies that the use of named entities as features slightly improves the performance of SVM-based stance detection. The initial and final versions of the dataset are shared with tweet identifiers and stance annotations only. In the current paper, we share and describe the final form of dataset in which annotations of both named entities and stance information are included. To the best of our knowledge, it is the only publicly-shared dataset annotated with named entities and stance, and we expect that it will help reveal (or confirm) the possible contribution of the use of named entities to the stance detection procedure.\\

The rest of the paper is organized as follows: In Section 2, related work on named entity recognition and stance detection is presented, with a particular emphasis on shared datasets. In Section 3, the details of the shared dataset, annotated both with stance and named entities, are provided, and finally Section 4 concludes the paper with a summary.

\section{Related Work}\label{sec:related}

NER is a long-studied problem and its most common definition is the extraction of named entities such as person, location, and organization names from texts, in addition to some temporal and numeric expressions \cite{chinchor1997muc}. Message Understanding Conference (MUC) series constitute an important milestone for NER research as they included NER competitions where related annotation guidelines, annotated datasets, and evaluation metrics were provided \cite{chinchor1997muc}. The focus of recent NER studies has shifted from well-formed texts to social media texts including mostly microblog posts, as demonstrated in studies such as \cite{ritter2011named,liu2013named}. Comprehensive surveys of NER research include \cite{nadeau2007survey,marrero2013named,goyal2018recent}.\\

Stance detection is a newly-emerging topic in NLP research, with an increasing number of on-topic studies conducted in recent years. It is usually defined as the automatic detection of whether the owner of a given piece of text is in favor of or against a given target. Therefore, a classification result as \emph{Favor}, \emph{Against}, or \emph{Neither\slash None} is generally expected at the end of the stance detection procedure. Similar to NER, earlier stance detection studies are performed on well-formed texts like online debate posts \cite{anand2011cats} and essays \cite{faulkner2014automated} and recent studies are mostly proposed for social media texts like tweets \cite{mohammad2017stance,mourad2018stance,kucuk_can_2018stance}.\\

Considering the limited number of stance detection datasets published so far, most of them are provided within the course of stance detection shared tasks at related conferences. To begin with, an annotated tweet dataset in English is presented and shared in SemEval-2016 shared task on stance detection in tweets \cite{mohammad2016semeval}. The details of this stance-annotated dataset is presented \cite{mohammad2016dataset} and its extended version also including sentiment annotations is described in \cite{mohammad2017stance}. In a similar shared task on stance detection, an annotated dataset of Chinese microblog texts is described \cite{xu2016chinesestancetask}. Finally, in \cite{taule2017stance_gender_ibereval}, a shared task on stance detection on Catalan and Spanish tweets is described, accompanied with the shared dataset used in the experiments. Apart from these datasets, which are used by several studies conducted after their publication, a tweet dataset annotated for multi-target stance detection is presented and publicly shared in \cite{sobhani2017dataset}.\\

The current paper describes and shares the named entity and stance annotations of the dataset used in \cite{kucuk_can_2018stance}. Within the course of this previous study, only stance annotations with tweet identifiers are publicly shared. The current study describes both the named entity and stance annotations of the same dataset. Hence it can be used both within research on NER and stance detection, particularly to determine possible effects of named entity annotations to stance detection procedure. The initial version of the dataset comprises 700 tweets in Turkish annotated for stance by a single native annotator \cite{kucuk2017stance}. This initial version is publicly shared at \url{https://github.com/dkucuk/Stance-Detection-Turkish-V1}. In the second version of the dataset, another native speaker also annotated the tweets for stance and hence the dataset includes 686 tweets (agreed by the two annotators) and the dataset is shared with stance annotations only, although named entities were also annotated (by a single annotator) and used during the reported stance detection experiments \cite{kucuk2017joint_stance_ner}. This second version of the dataset is available at \url{https://github.com/dkucuk/Stance-Detection-Turkish-V2}. Finally, in \cite{kucuk_can_2018stance}, the total number of tweets in the dataset is increased to 1,065 as tagged with stance information and agreed by the two annotators, and this final version of the dataset, with stance information only, is made available at \url{https://github.com/dkucuk/Stance-Detection-Turkish-V3}.

\section{Named Entity and Stance Annotated Tweet Dataset}\label{sec:desc}

As mentioned in the previous section, our tweet dataset has 1,065 tweets in Turkish and are about two sports clubs in Turkey, namely \emph{Galatasaray} and \emph{Fenerbah\c{c}e}, which are the stance targets used during the annotations. The stance classes are \emph{Favor} and \emph{Against}. For 537 tweets, the stance target is \emph{Galatasaray} (269 annotated as \emph{Favor} and 268 annotated as \emph{Against}), and for 528 tweets, the stance target is \emph{Fenerbah\c{c}e} (269 annotated as \emph{Favor} and 259 annotated as \emph{Against}).\\

Regarding the named entity annotations in the dataset, the named entity classes considered are \emph{Person}, \emph{Location}, and \emph{Organization}, since only these three classes of named entities are considered during the annotation procedure. There are a total of 1,879 named entities annotated in the whole dataset, where 868 entities are annotated in the 537 tweets with the stance target \emph{Galatasaray}, and the remaining 1,011 entities are observed and annotated in the 528 tweets associated the stance target \emph{Fenerbah\c{c}e}.\\

We should note that, while the stance annotations on the dataset are performed by two annotators, the named entity annotations are performed by a single annotator only.\\

The ultimate form of the annotated dataset is also made publicly available at \url{https://github.com/dkucuk/Tweet-Dataset-NER-SD} as a comma-separated values (CSV) file. It includes the following items for each tweet and each item is separated from the others with semicolons.
\begin{itemize}
    \item   The tweet identifier
    \item   Stance target (as \emph{Galatasaray} or \emph{Fenerbah\c{c}e})
    \item   Stance class (as \emph{Favor} or \emph{Against})
    \item   A set of triplets where each triplet corresponds to a single named entity annotation and triplets are similarly separated from each other by semicolons. The three elements in each item are separated from each other by commas. The first element in each triplet is the index of the starting character of the named entity (not counting the whitespace characters and punctuation marks), the second element is the index of the first character after the ending character of this named entity, and the final element of the triplet is the named entity type (as PERSON, LOCATION, or ORGANIZATION). The starting index of such considered characters is 0, i.e., zero-based indexing is employed. This annotation scheme for named entities in tweets has been previously employed in \cite{kuccuk2016named}.
\end{itemize}

The following four sample tweets from the dataset are excerpted from the aforementioned preceding study \cite{kucuk_can_2018stance}, this time also accompanied with their annotated versions with named entities (NEs) and the corresponding rows in the dataset covering both stance and named entity annotations.\\

Original Tweet in Turkish: \texttt{ve biz iyi ki Galatasarayl{\i}y{\i}z}\\
English Translation: \texttt{and fortunately we are supporters of Galatasaray}\\
Stance Target: \texttt{Galatasaray}\\
Stance Annotation: \texttt{Favor}\\
NE-Annotated Version: \texttt{ve biz iyi ki <ENAMEX TYPE="ORGANIZATION">Galatasaray</ENAMEX>{\i}y{\i}z}\\
Corresponding Row in Dataset: \texttt{633686096602337280;Galatasaray;Favor;10,21,ORGANIZATION}\\

Original Tweet in Turkish: \texttt{Bu grup ha\c{s}lar Galatasaray{\i} :D}\\
English Translation: \texttt{This group will boil Galatasaray :D}\\
Stance Target: \texttt{Galatasaray}\\
Stance Annotation: \texttt{Against}\\
NE-Annotated Version: \texttt{Bu grup ha\c{s}lar <ENAMEX TYPE="ORGANIZATION">Galatasaray</ENAMEX>{\i}}\\
Corresponding Row in Dataset: 6\texttt{36939237473034240;Galatasaray;Against;12,23,ORGANIZATION}\\

Original Tweet in Turkish: \texttt{Fenerbah\c{c}eli olmak ayr{\i}cal{\i}kt{\i}r...}\\
English Translation: \texttt{It is a privilege to be the supporter of Fenerbah\c{c}e...}\\
Stance Target: \texttt{Fenerbah\c{c}e}\\
Stance Annotation: \texttt{Favor}\\
NE-Annotated Version: \texttt{<ENAMEX TYPE="ORGANIZATION">Fenerbah\c{c}e</ENAMEX>li olmak\\ ayr{\i}cal{\i}kt{\i}r...}\\
Corresponding Row in Dataset: \texttt{635802060638801920;Fenerbah\c{c}e;Favor;0,10,ORGANIZATION}\\

Original Tweet in Turkish: \texttt{Kanser olmaya haz{\i}r m{\i}y{\i}z ? \#Fenerinma\c{c}{\i}var}\\
English Translation: \texttt{Ready to get cancer ? \#Fenerhasamatch}\\
Stance Target: \texttt{Fenerbah\c{c}e}\\
Stance Annotation: \texttt{Against}\\
NE-Annotated Version: \texttt{Kanser olmaya haz{\i}r m{\i}y{\i}z ?\\ \#<ENAMEX TYPE="ORGANIZATION">Fener</ENAMEX>inma\c{c}{\i}var}\\
Corresponding Row in Dataset: \texttt{635476961821913089;Fenerbah\c{c}e;Against;22,27,ORGANIZATION}\\

This annotated dataset is a significant resource particularly due to the below listed points:
\begin{itemize}
    \item   To the best of our knowledge, this is the first publicly-shared tweet dataset annotated both for stance and included named entities. Although the use of named entities to improve stance detection has been considered in our preceding studies \cite{kucuk2017joint_stance_ner,kucuk_can_2018stance}, within the course of these studies only stance annotations are publicly shared. Hence, it can be stated that the current paper is built upon and complements the work reported in \cite{kucuk2017joint_stance_ner,kucuk_can_2018stance}, where the latter two studies also present the evaluation results of stance detection experiments using named entities as features.
    \item   The language of the dataset is Turkish which is still a resource-scarce language and hence we believe that this language resource will help increase the number of NER and social media analysis studies on Turkish content.
\end{itemize}

\section{Conclusion}\label{sec:conc}

Publicly-shared language resources like annotated datasets are critical for the development of high-performance approaches and systems for different NLP problems. These resources can be used both for the training of supervised learning algorithms and for the evaluation and comparison of different solution proposals. In this paper, we present a tweet dataset in Turkish which is annotated both for NER and stance detection. NER is an NLP problem studied for a long time while stance detection is a comparatively new topic. To the best of our knowledge, this is the first such dataset including both named entity and stance annotations, and hence, can be used in both NER and stance detection research. More importantly, it can be used to reveal the possible contribution of the annotations for one of these problems to the solution of the second one. Although previous research reports results indicating that named entities can be improving features for stance detection, further related experiments with different types of classifiers will be beneficial to support this relationship between these two NLP tasks.

\bibliographystyle{lrec2006}

\end{document}